\documentclass[a4paper]{article}  

\usepackage[textwidth=17cm]{geometry}
\usepackage{amsmath,amsfonts,amssymb}
\usepackage{graphicx}
\usepackage{multirow}
\usepackage{multicol}
\usepackage{booktabs} 
\usepackage[colorlinks=true, allcolors=blue]{hyperref}

\title{BVI-Mamba: Video Enhancement Using a Visual State-Space Model for Low-Light and Underwater Environments}

\author{Guoxi~Huang, Ruirui~Lin, Yini~Li, David~R.~Bull, \\ and Nantheera~Anantrasirichai \\ Visual Information Laboratory, \\ University of Bristol, Bristol, UK}

\begin{document} 
\maketitle

\begin{abstract}
Videos captured in low-light and underwater conditions often suffer from distortions such as noise, low contrast, color imbalance, and blur. These issues not only limit visibility but also degrade automatic tasks like detection. Post-processing is typically required but can be time-consuming. AI-based tools for video enhancement also demand significantly more computational resources compared to image-based methods. This paper introduces a novel framework, Visual Mamba, designed to reduce memory usage and computational time by leveraging the Visual State Space (VSS) model. The framework consists of two modules: (i) a feature alignment module, where spatio-temporal displacement between input frames is registered in the feature space, and (ii) an enhancement module, where noise removal and brightness adjustment are performed using a UNet-like architecture, with all convolutional layers replaced by VSS blocks. Experimental results show that the Visual Mamba technique outperforms Transformer and convolution-based models in both low-light and underwater video enhancement tasks. Code is available on line at \url{https://github.com/russellllaputa/BVI-Mamba}.
\end{abstract}

\section{INTRODUCTION}
\label{sec:intro}  

Images and videos are crucial for capturing and conveying information, yet their quality often suffers from distortions and noise, especially in challenging environments such as low-light or underwater settings. Inadequate exposure settings, characterized by imbalances in the exposure triangle (shutter speed, aperture, ISO), often result in a low signal-to-noise ratio, noise from elevated ISO levels, blurring from slow shutter speeds, and focus issues due to a shallow depth of field. Underwater imaging introduces further complexities due to the dynamic nature of water and its particulate content. Scattering and absorption of light in these conditions cause blurring, halo effects, color distortions, and reduced contrast. Such quality degradations not only compromise the aesthetic integrity of images but also adversely affect the performance of computer vision tasks, including object detection, classification, and tracking in critical applications such as surveillance systems and autonomous vehicle navigation.

In challenging environments characterized by low contrast and susceptibility to blurring from long exposures, significant technological advancements have been made to enhance image and video quality. AI-based methods, particularly those leveraging deep learning, have emerged as the state-of-the-art for improving content captured in low-light conditions. For instance, LEDNet \cite{Zhou:LEDNet:2022} utilizes a novel activation function specifically designed to amplify features in dimly lit scenes. SNR-Aware \cite{Xu:SNR:2022} enhances image quality by estimating spatially varying SNR maps that allow for more refined image processing. Additionally, NeRCo \cite{Yang:Implicit:2023} tackles the inherent difficulties of low-light photography using Implicit Neural Representation, while Diff-Retinex \cite{Yi:Diff:2023} applies generative networks based on Retinex theory to simulate conditions akin to normal lighting.

Recent advancements also extensively employ convolutional neural networks (CNNs), transformers, and diffusion models, each presenting a unique balance between performance enhancement and computational efficiency. Inspired by the Mamba framework \cite{gu2023mamba} and its subsequent adaptations for vision tasks \cite{liu2024vmamba,zhu2024vision}, which are noted for their linear computational complexity in processing long sequences, we introduce BVI-Mamba for low-light video enhancement (LLVE). This approach is geared towards optimizing video quality in low-light conditions while managing the computational demands effectively.

We introduce BVI-Mamba, an innovative solution for low-light video enhancement, tailored for challenging environments, including low-light and underwater scenarios. This tool adapts the framework of STA-SUNet~\cite{Lin:ICIP:2024} but innovates by substituting all Swin Transformer blocks~\cite{Liang:SwinIR:2021} in the reconstruction module with State Space Models (SSMs) featuring a 2D Selective Scan (SS2D) module~\cite{liu2024vmamba}. 
BVI-Mamba begins its process by aligning multiple input frames at the spatio-temporal feature level using a Pyramid, Cascading, and Deformable (PCD) alignment integrated with Temporal and Spatial Attention (TSA) fusion modules, as delineated in Ref. \cite{Wang:EDVR:2019}. This sophisticated alignment module adeptly extracts and processes temporal and spatial details, ensuring the aligned features in consecutive frames maintain temporal consistency and mitigate issues such as flickering or brightness fluctuations.
For enhancement, BVI-Mamba employs a U-Net-like architecture, replacing traditional convolutional layers with Visual State Space (VSS) blocks, as proposed in the Visual Mamba model \cite{gu2023mamba}. This adjustment overcomes the limitations of standard convolutional approaches, which often struggle with larger patches and long-range dependencies due to their content-dependent nature. By utilizing VSS blocks, BVI-Mamba captures both local details and global features more effectively, resulting in a model that is not only more efficient but also superior in handling complex video enhancement tasks compared to conventional vision transformers.



\section{Existing methods for low-light and underwater video enhancement}

\subsection{Low-light video enhancement}

Recent advancements in deep learning technologies have significantly propelled the field of image enhancement forward~\cite{retinexDIP,LECARM}. However, video enhancement techniques that leverage machine learning are still developing~\cite{Li:Low:2022}, challenged by factors such as a greater number of unknown parameters, dataset limitations, computational demands, and high memory requirements.

Typically, to process the video, multiple input frames are used in these processes. SMOID \cite{Jiang:learn:2019} adapted the UNet architecture \cite{ronneberger2015unet} to process multiple frames, although this approach struggles with fast-moving objects. To address this, several techniques aim to align feature maps of neighboring frames with the current frame~\cite{zhou2021rta}. Initial approaches utilized recurrent neural networks~\cite{Wang:enhancing:2019}, but more recently, Deformable Convolution Networks (DCN)\cite{Dai:Deformable:2017} have been favored for their effectiveness in this context. An iterative re-weighting of features helps to reduce errors in merging\cite{zhou2021rta}. Liu et al.\cite{Liu:low:2023} introduced a method to adaptively combine frames using synthetic events from videos, based on the movement speeds of objects. Fu et al. \cite{Fu:dancing:2023}, creators of the DID dataset, developed a light-adjustable network inspired by Retinex theory, though their implementation is not available publicly. Recently, a method based on a wavelet conditional diffusion model \cite{Lin:CVMP:2024} was proposed, demonstrating superior performance compared to CNNs and transformer-based methods.
Some approaches depend on Bayer raw video inputs, which are not practical for consumer-grade cameras~\cite{Chen_2019_ICCV, Jiang:learn:2019, triantafyllidou2020low}. Given the scarcity of paired training datasets, some researchers have explored unpaired training methods, such as using CycleGAN \cite{anantrasirichai:Contextual:2021}. This technique has also been applied to convert low-light RAW videos into RGB format \cite{triantafyllidou2020low}.

Applying Mamba to low-light enhancement is still in its early stages and has only been applied to images, not videos. Examples include Retinexmamba \cite{bai2024retinexmamba}, Wave-Mamba \cite{Zou:WaveMamba:2024}, and FourierTMamba \cite{Peng:FourierTMamba:2024}.

\subsection{Underwater video enhancement}

Research on underwater image enhancement is well-established, yet few studies tackle the challenges associated with enhancing underwater video quality. One primary obstacle is the scarcity of datasets necessary for training learning-based methods \cite{Xie_2024_CVPR}. Models have incorporated physical properties, including the characterization of underwater light propagation \cite{dekker2001imaging}, radiative transfer theory \cite{dutre2018advanced}, and light reflectance models \cite{dekker2001imaging}. The most commonly used model is the Revised Underwater Image Formation Model (UIFM) \cite{akkaynak2018revised}, which is widely recognized for its distinct modeling of attenuation and backscattering factors for direct and backscattered signals \cite{zhang2024atlantis, Wang2025}. However, the reliance on single-image processing often leads to inconsistent results in similar scenarios, which diminishes the reliability and effectiveness of these methods.

 \begin{figure} [t]
   \begin{center}
    \includegraphics[width=\textwidth]{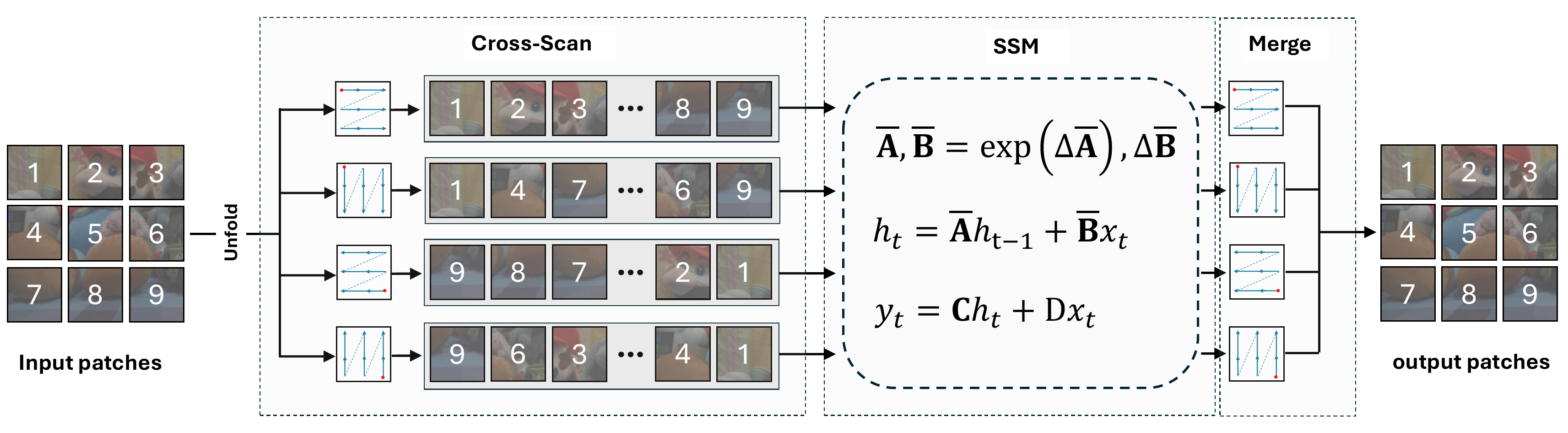}
   \end{center}
    \caption{Diagram of 2D-Selective-Scan (SS2D). SS2D contains three steps, including cross-scan, SSM processing and the merge operation.}
    \label{fig: ss2d}
\end{figure}

\section{Preliminaries}
\subsection{Revisiting State Space Model (SSM)}
A classical SSM is a continuous linear time-invariant (LTI) system that maps a 1-dimensional input $x(t) \in \mathbb{R}$ to an output $y(t) \in \mathbb{R}$ via an implicit latent state $h(t) \in \mathbb{R}^d$, which can be expressed by the following linear ordinary differential equations (ODE):

\begin{equation}
\begin{aligned}
h'(t) =\mathbf{A} h(t-1)+\mathbf{B} x(t)&, ~~~~~~ \mathbf{A} \in \mathbb{R}^{d \times d}, \mathbf{B} \in \mathbb{R}^{d \times 1},  \\
y(t) =\mathbf{C} h(t)+ D x(t)&, ~~~~~~ \mathbf{C} \in \mathbb{R}^{1 \times d}, D \in \mathbb{R}
\end{aligned}
\label{eq: conti_ssm}
\end{equation}
In order to integrate SSM into deep learning-based models, the continuous parameters, $\mathbf{A}$ and $ \mathbf{B}$, are discretized using the zero-order hold. Specifically, the discretized parameters $\overline{\mathbf{A}}$ and $\overline{\mathbf{B}}$ are formulated by:
\begin{equation}
\begin{aligned}
& \overline{\mathbf{A}}=\exp (\Delta \mathbf{A}) \\
& \overline{\mathbf{B}}=(\Delta \mathbf{A})^{-1}(\exp (\mathbf{A})-\mathbf{I}) \cdot \Delta \mathbf{B} \approx \Delta \mathbf{B},
\end{aligned}
\end{equation}
where $\Delta \in \mathbb{R} $ is a timescale parameter.
Thereafter, we can rewrite \eqref{eq: conti_ssm} in a discretized format:

\begin{equation}
\begin{aligned}
h_t & =\overline{\mathbf{A}} h_{t-1}+\overline{\mathbf{B}} x_t \\
y_t & =\mathbf{C} h_t+\mathbf{D} x_t
\end{aligned}
\label{eq: discre_ssm}
\end{equation}
With \eqref{eq: discre_ssm}, the discretized SSM can either be computed as a recurrence or global convolution. 

\subsection{Mamba} Mamba~\cite{gu2023mamba}, the cutting-edge SMM-based technique, is considered a linear Transformer, characterized by its linear complexity in attention modeling. The authors introduce a \textbf{selective scan} method by parameterizing the SSM parameters, $\textbf{B}, \textbf{C}, \Delta$, based on the input. By integrating with the selection mechanism, the SSM is endured with the context-awareness, which is hypothesized to be a core capability for in-context learning in LLMs~\cite{olsson2022context}.
As convolutions do not accommodate dynamic weights, Gu et al.~\cite{gu2023mamba} proposed to parallelly compute hidden state $h_{t-1}$ in Eq.\eqref{eq: discre_ssm} by employing parallel prefix scan~\cite{blelloch1990prefix}.

\section{Proposed method -- BVI-MAMBA}

Building upon the Mamba framework, we propose BVI-Mamba, a novel approach for low-light video enhancement. The method operates on discrete, non-overlapping patches of video frames, which are subsequently arranged sequentially through a structured scanning strategy.

\textbf{2D-Selective-Scan (SS2D).} To apply the selective scan~\cite{gu2023mamba} to visual data, Liu et. al~\cite{liu2024vmamba} propose a 2D-Selective-Scan module. As shown in Fig.~\ref{fig: ss2d}, the SS2D module enables 2D structure awareness by processing a video frame from four distinct directions. It first unfolds the input patches into sequences along four traversal paths (i.e., Left-Right, Top-Down, Right-Left, Down-Top). Each sequence is subsequently processed by SSM. Afterwards, the four processed sequences are merged into one, forming the output patches.


 \begin{figure} [t]
   \begin{center}
    \includegraphics[width=\textwidth]{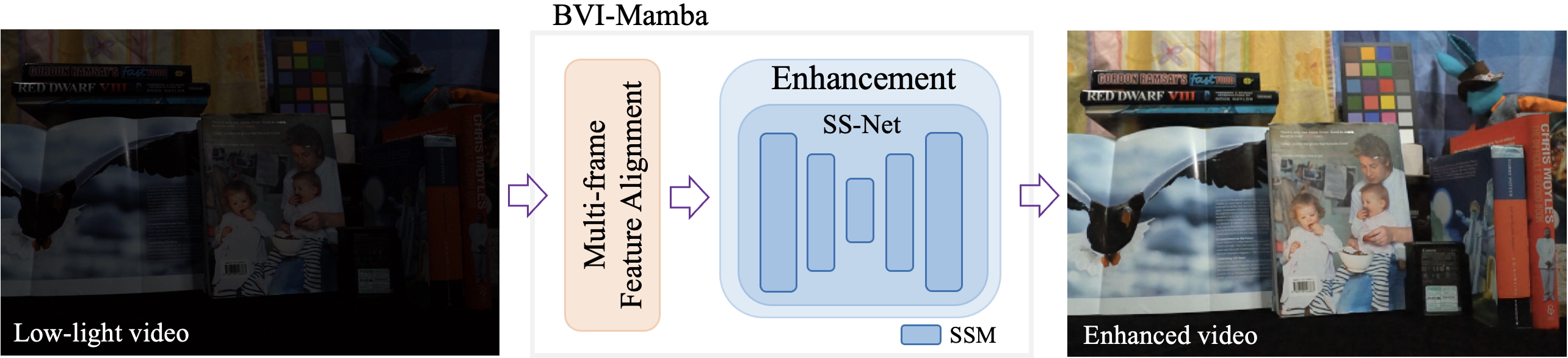}
   \end{center}
    \caption{Diagram of the proposed BVI-Mamba for video enhancement.}
    \label{fig:diagram}
\end{figure}

In the development of the proposed BVI-Mamba, we incorporate the STA-SUNet framework \cite{Lin:ICIP:2024}. Our approach utilizes a two-stage framework that combines a multi-frame feature alignment module with a UNet-based enhancement module, as shown in Fig. \ref{fig:diagram}. Specifically, we employ the Pyramid, Cascading, and Deformable (PCD) alignment module from EDVR \cite{Wang:EDVR:2019}, omitting the feature fusion module for a more streamlined design. This modified setup results in a significantly lighter framework compared to the conventional EDVR approach. For the enhancement stage, we adopt a U-Net-like structure where standard convolutions are substituted with VSS blocks. These blocks are designed to effectively capture both local and global features within a hierarchical architecture.

The formulation of VSS block~\cite{liu2024vmamba} in layer $l$ can be expressed as
\begin{equation}
    \begin{aligned}
        & \textbf{h}_l = \textrm{SS2D} \left(\textrm{LN}\left(\textbf{h}_{l-1} \right)  \right) + \textbf{h}_{l-1},\\
        & \textbf{h}_{l+1} = \textrm{FFN}\left({\textrm{LN}\left(\textbf{h}_l \right)}\right ) + \textbf{h}_l,
    \end{aligned}
\end{equation}
where FFN denotes the feedforward network and LN denotes layer normalisation. $\textbf{h}_{l-1}$ and $\textbf{h}_l$ denote the input and output in the $l$-th layer, respectively. 



\section{EXPERIMENTAL RESULTS}
\label{sec:results}

\subsection{Datasets}
We assessed our method's effectiveness using two datasets specifically designed for benchmarking low-light video enhancement: DID \cite{Fu:dancing:2023}, and BVI-RLV \cite{Lin:BVIRLV:2024}, and one underwater dataset, BVI-Coral, as used in \cite{Gough2025AquaNeRF}.  The DID dataset consists of 413 video pairs filmed using five different cameras, with dynamic scenes created via an electric gimbal and brightness adjustments made with an ND filter. However, this dataset also suffers from misalignment issues and occasional lens dirt, which could affect video quality assessments. The BVI-RLV dataset offers fully registered HD video pairs in both normal and low light conditions, with 40 scenes at two lighting levels, 10\% and 20\%. It includes 32 training scenes and 8 for testing, with videos ranging from 200 to 700 frames. BVI-Coral provides 22 underwater videos intended for underwater 3D reconstruction, but we tested only the videos captured in the deep sea that appear to be in low light conditions.

\subsection{Experiment setting}

We assessed the performance of the proposed Mamba-based method and compared it with three different architectures: PCDUNet~\cite{BVI-Lowlight}, a CNN-based method; STA-SUNet \cite{Lin:ICIP:2024}, a Transformer-based method; and BVI-CDM \cite{Lin:CVMP:2024}, a diffusion model-based method. For training and evaluation splits, we adhered to the ratios specified in the original papers: 73:27 for DRV and 80:20 for BVI-RLV. We aligned training parameters with those in the original publications to ensure consistency and fairness in our comparisons. Where training details were absent, we defaulted to 200k iterations per model. For underwater video testing, as ground truth of underwater vidoes is not available, we employed the model trained with BVI-RLV dataset and tested on the BVI-Coral dataset. We measure the visual quality of the enhanced videos using the objective metrics, including Peak signal-to-noise ratio (PSNR), Structural similarity index measure (SSIM), and Perceptual similarity like LPIPS \cite{zhang2018perceptual}.

The proposed framework was implemented in Python, leveraging PyTorch and CUDA. A deformable convolution component, developed in C++ in Ref. \cite{Wang:EDVR:2019}, was integrated into the model. All training and testing were performed on a high-performance computing system equipped with Nvidia P100 GPUs. The system comprises 32 GPU nodes, each containing two cards, and a single GPU login node, providing access to a total of 65 P100 GPUs.

\subsection{Performance}

\subsubsection{Low-light videos}

\begin{figure} [t]
   \begin{center}
    \includegraphics[width=\textwidth]{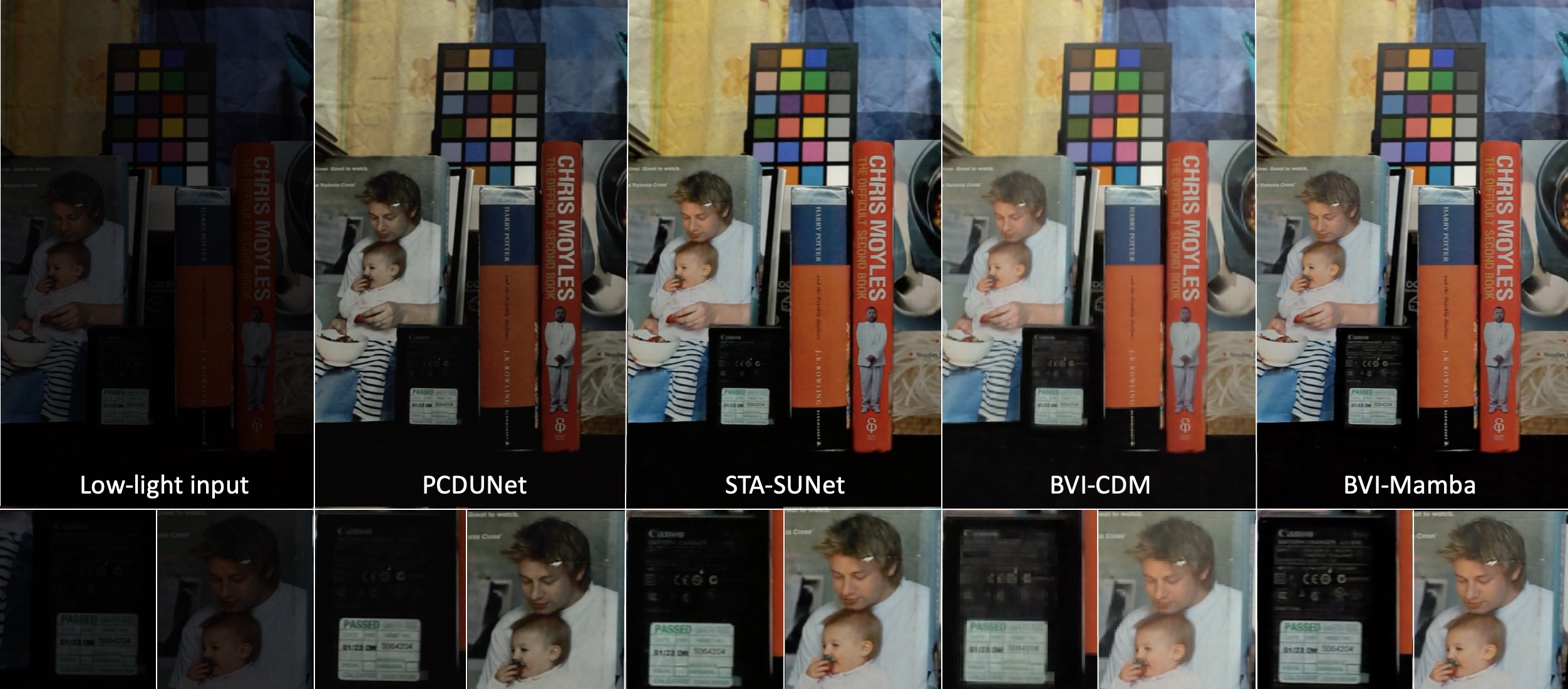}
   \end{center}
    \caption{Low-light enhancement comparison (left to right): the original frame, results of PCDUNet, STA-SUNet, BVI-CDM and BVI-Mamba}
    \label{fig:lowlightresults}
\end{figure}

We present experimental results in Table~\ref{tab: more_results_mamba}, comparing our approach with other benchmarking methods. Our results demonstrate that the visual Mamba technique outperforms Transformer- and convolution-based counterparts in LLVE tasks. Additionally, subjective assessments of the enhanced outputs are depicted in Fig.~\ref{fig:lowlightresults}. These visual results illustrate that BVI-Mamba not only effectively reduces noise and sharpens images but also adeptly preserves the fine details within the scenes. .

\begin{table}[t]
    \caption{Results of BVI-Mamba compared with other benchmarking methods when trained on the BVI-RLV and DRV datasets. The reported results are averaged across the test datasets.}
    \label{tab: more_results_mamba}
    \begin{center}  
    \begin{tabular}{l|ccc|ccc}
    \toprule
       \multirow{2}{*}{Method} & \multicolumn{3}{c|}{DRV \cite{Chen_2019_ICCV}} & \multicolumn{3}{c}{BVI-RLV \cite{Lin:BVIRLV:2024}} \\ 
        \cmidrule(r){2-7} 
        & PSNR & SSIM & LPIPS & PSNR & SSIM & LPIPS \\ \hline
        PCDUNet (CNN-based) & 15.00 & 0.492 & 0.422 & 19.50 & 0.757 & 0.316 \\
        STA-SUNet (Transformer-based) & 14.19 & 0.424 & 0.424 & 20.64 & 0.765 & 0.243 \\
        BVI-CDM (Diffusion model-based) & 19.56 & 0.575 & 0.350 & 22.20 & 0.773 & 0.175 \\ \hline
        BVI-Mamba & \textbf{22.95} & \textbf{0.784} & \textbf{0.195} & \textbf{22.93} & \textbf{0.820} & \textbf{0.146} \\
    \bottomrule
\end{tabular}
\end{center}
\end{table}

\subsubsection{Low-light underwater videos}
\begin{figure} [t]
   \begin{center}
    \includegraphics[width=\textwidth]{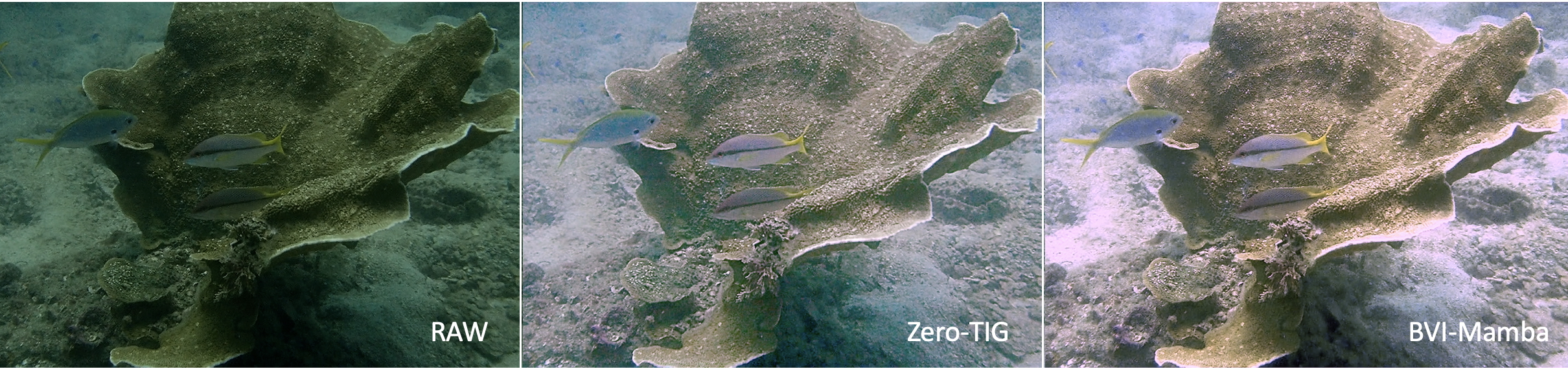}
   \end{center}
    \caption{Underwater enhancement comparison (left to right): the original frame, results of Zero-TIG, and BVI-Mamba}
    \label{fig:underwaterresults}
\end{figure}

Underwater video data rarely has ground truth available. Therefore, we employ a pretrained model generated using the BVI-RLV dataset. However, underwater scenes suffer from severe color imbalance due to the absorption and scattering of light in water. To address this, we apply color balance adjustment using chromatic adaptation\footnote{\url{http://www.brucelindbloom.com/index.html?Eqn_ChromAdapt.html}} before feeding the data into the enhancement model. The scene illuminant is estimated simply from the mean of the red, green, and blue channels. Since no ground truth is available, we assess the method using subjective evaluation and compare the result with that from Zero-TIG~\cite{Zero-TIG:2025}, as shown in Fig. \ref{fig:underwaterresults}.

\section{CONCLUSION}
\label{sec:conclude}

This paper introduces BVI-Mamba, a novel Mamba-based framework for low-light video enhancement that leverages a visual state space approach. The core architecture of BVI-Mamba consists of a frame alignment module paired with a multi-scale enhancement module, designed to effectively address the challenges associated with low-light video content. We trained the BVI-Mamba model on a dataset that was fully registered and contained a wide range of motions and subsequently evaluated its efficacy on various low-light datasets. Our comparative analysis includes several leading models in the domain. Quantitative evaluations, focusing on PSNR and SSIM, alongside subjective assessments, show that BVI-Mamba significantly outperforms existing methods. This advancement not only demonstrates the robustness of the Mamba-based approach in handling diverse and challenging video enhancement tasks but also establishes BVI-Mamba as a superior solution for improving video quality in low-light conditions.

\section*{acknowledgments} 
 
This work was supported by the UKRI MyWorld Strength in Places Programme (SIPF00006/1) and the EPSRC ECR International Collaboration Grants (EP/Y002490/1).

\bibliography{report} 

@String(CVPR= {IEEE Conf. Comput. Vis. Pattern Recog.})

@String(ICCV= {Int. Conf. Comput. Vis.})

@String(ECCV= {Eur. Conf. Comput. Vis.})

@String(ICIP = {IEEE Int. Conf. Image Process.})

@String(AAAI = {AAAI})

@String(CVPR  = {CVPR})

@String(ICCV  = {ICCV})

@String(ECCV  = {ECCV})

@String(TCSVT = {IEEE TCSVT})

@String(ICIP  = {ICIP})

@INPROCEEDINGS{anantrasirichai:Contextual:2021,  author={Anantrasirichai, N. and Bull, David},  booktitle={ICIP proc.},   title={Contextual Colorization and Denoising for Low-Light Ultra High Resolution Sequences},   year={2021},  volume={},  number={},  pages={1614-1618}}

@inproceedings{ronneberger2015unet,
  title     = {U-Net: Convolutional Networks for Biomedical Image Segmentation},
  author    = {Ronneberger, Olaf and Fischer, Philipp and Brox, Thomas},
  booktitle = {Medical Image Computing and Computer-Assisted Intervention (MICCAI)},
  year      = {2015}
}

@INPROCEEDINGS{Liang:SwinIR:2021,
  author={Liang, Jingyun and Cao, Jiezhang and Sun, Guolei and Zhang, Kai and Van Gool, Luc and Timofte, Radu},
  booktitle={2021 IEEE/CVF International Conference on Computer Vision Workshops (ICCVW)}, 
  title={SwinIR: Image Restoration Using Swin Transformer}, 
  year={2021},
  volume={},
  number={},
  pages={1833-1844},
  doi={10.1109/ICCVW54120.2021.00210}}

@INPROCEEDINGS{Jiang:learn:2019,  author={Jiang, Haiyang and Zheng, Yinqiang},  booktitle={2019 IEEE/CVF International Conference on Computer Vision (ICCV)},   title={Learning to See Moving Objects in the Dark},   year={2019},  volume={},  number={},  pages={7323-7332},  doi={10.1109/ICCV.2019.00742}}

@INPROCEEDINGS{Dai:Deformable:2017,
author={J. {Dai} and H. {Qi} and Y. {Xiong} and Y. {Li} and G. {Zhang} and H. {Hu} and Y. {Wei}},
booktitle={ICCV},
title={Deformable Convolutional Networks},
year={2017},
volume={},
number={},
pages={764-773},
doi={10.1109/ICCV.2017.89},
ISSN={2380-7504},
month={Oct},}

@inproceedings{Gough2025AquaNeRF,
  author    = {Luca Gough and Adrian Azzarelli and Fan Zhang and Nantheera Anantrasirichai},
  title     = {{AquaNeRF}: Neural Radiance Fields in Underwater Media with Distractor Removal},
  booktitle = {IEEE International Symposium on Circuits and Systems},
  year      = {2025}
}

@article{Zero-TIG:2025,
author    = {Yini Li and Nantheera Anantrasirichai},
  title     = {{Zero-TIG: Temporal} Consistency-Aware Zero-Shot Illumination-Guided Low-light Video Enhancement},
  year      = {2025},
  journal   = {arXiv:2503.11175},
}

@article{BVI-Lowlight,
  author    = {Nantheera Anantrasirichai and Ruirui Lin and Anna Malyugina and David Bull},
  title     = {{BVI-Lowlight: Fully} registered benchmark dataset for low-light video enhancement},
  year      = {2024},
  journal   = {arXiv:2402.01970},
  url       = {https://arxiv.org/abs/2402.01970}
}

@inproceedings{zhang2018perceptual,
  title={The Unreasonable Effectiveness of Deep Features as a Perceptual Metric},
  author={Zhang, Richard and Isola, Phillip and Efros, Alexei A and Shechtman, Eli and Wang, Oliver},
  booktitle={CVPR},
  year={2018}
}

@INPROCEEDINGS{Wang:enhancing:2019,
  author={Wang, Wei and Chen, Xin and Yang, Cheng and Li, Xiang and Hu, Xuemei and Yue, Tao},
  booktitle={2019 IEEE/CVF International Conference on Computer Vision (ICCV)}, 
  title={Enhancing Low Light Videos by Exploring High Sensitivity Camera Noise}, 
  year={2019},
  volume={},
  number={},
  pages={4110-4118},
  doi={10.1109/ICCV.2019.00421}}

@ARTICLE{Li:Low:2022,
  author={Li, Chongyi and Guo, Chunle and Han, Linghao and Jiang, Jun and Cheng, Ming-Ming and Gu, Jinwei and Loy, Chen Change},
  journal={IEEE Transactions on Pattern Analysis and Machine Intelligence}, 
  title={Low-Light Image and Video Enhancement Using Deep Learning: A Survey}, 
  year={2022},
  volume={44},
  number={12},
  pages={9396-9416},
  doi={10.1109/TPAMI.2021.3126387}}

@article{Liu:low:2023, title={Low-Light Video Enhancement with Synthetic Event Guidance}, volume={37}, DOI={10.1609/aaai.v37i2.25257},  number={2}, journal={Proceedings of the AAAI Conference on Artificial Intelligence}, author={Liu, Lin and An, Junfeng and Liu, Jianzhuang and Yuan, Shanxin and Chen, Xiangyu and Zhou, Wengang and Li, Houqiang and Wang, Yan Feng and Tian, Qi}, year={2023}, month={Jun.}, pages={1692-1700} }

@inproceedings{triantafyllidou2020low,
  title={Low Light Video Enhancement using Synthetic Data Produced with an Intermediate Domain Mapping},
  author={Triantafyllidou, Danai and Moran, Sean and McDonagh, Steven and Parisot, Sarah and Slabaugh, Gregory},
  booktitle={European Conference on Computer Vision},
  pages={103--119},
  year={2020},
  organization={Springer}
}

@inproceedings{Fu:dancing:2023,
    author = {Huiyuan Fu and Wenkai Zheng and Xicong Wang and Jiaxuan Wang and Heng Zhang and Huadong Ma},
    title = {Dancing in the Dark: {A} Benchmark towards General Low-light Video Enhancement},
    booktitle = {Proceedings of the IEEE/CVF International Conference on Computer Vision (ICCV)},
    year = {2023}
}

@inproceedings{zhou2021rta, 
  title={Revisiting Temporal Alignment for Video Restoration},
  author={Kun Zhou and Wenbo Li and Liying Lu and Xiaoguang Han and Jiangbo Lu}, 
  booktitle={Proceedings of the IEEE/CVF Conference on Computer Vision and Pattern Recognition},
  year={2022} 
}

@InProceedings{Wang:EDVR:2019,
          author = {Wang, Xintao and Chan, Kelvin C.K. and Yu, Ke and Dong, Chao and Loy, Chen Change},
          title = {{EDVR: V}ideo Restoration with Enhanced Deformable Convolutional Networks},
          booktitle = {The IEEE Conference on Computer Vision and Pattern Recognition (CVPR) Workshops},
          month = {June},
          year = {2019}
          }

@InProceedings{Chen_2019_ICCV,
author = {Chen, Chen and Chen, Qifeng and Do, Minh N. and Koltun, Vladlen},
title = {Seeing Motion in the Dark},
booktitle = {Proceedings of the IEEE/CVF International Conference on Computer Vision (ICCV)},
month = {October},
year = {2019}
}

@INPROCEEDINGS{Lin:ICIP:2024,  author={Ruirui Lin and Nantheera Anantrasirichai and Alexandra Malyugina and David Bull},  booktitle={IEEE International Conference on Image Processing},   title={A Spatio-Temporal Aligned SUNet Model FOR LOW-LIGHT VIDEO ENHANCEMENT},   year={2024},  volume={},  number={},  pages={}}

@INPROCEEDINGS{gu2023mamba,
  title={{Mamba: Linear}-time sequence modeling with selective state spaces},
  author={Gu, Albert and Dao, Tri},
  booktitle={Conference on Language Modeling},
  year={2024}
}

@article{liu2024vmamba,
  title={{Vmamba: Visual} state space model},
  author={Liu, Yue and Tian, Yunjie and Zhao, Yuzhong and Yu, Hongtian and Xie, Lingxi and Wang, Yaowei and Ye, Qixiang and Liu, Yunfan},
  journal={arXiv preprint arXiv:2401.10166},
  year={2024}
}

@INPROCEEDINGS{Xu:SNR:2022,
  author={Xu, Xiaogang and Wang, Ruixing and Fu, Chi-Wing and Jia, Jiaya},
  booktitle={2022 IEEE/CVF Conference on Computer Vision and Pattern Recognition (CVPR)}, 
  title={SNR-Aware Low-light Image Enhancement}, 
  year={2022},
  volume={},
  number={},
  pages={17693-17703},
  doi={10.1109/CVPR52688.2022.01719}}

@InProceedings{Zhou:LEDNet:2022,
author="Zhou, Shangchen
and Li, Chongyi
and Change Loy, Chen",
editor="Avidan, Shai
and Brostow, Gabriel
and Ciss{\'e}, Moustapha
and Farinella, Giovanni Maria
and Hassner, Tal",
title="LEDNet: Joint Low-Light Enhancement and Deblurring in the Dark",
booktitle="Computer Vision -- ECCV 2022",
year="2022",
pages="573--589",
}

@inproceedings{Zou:WaveMamba:2024,
author = {Zou, Wenbin and Gao, Hongxia and Yang, Weipeng and Liu, Tongtong},
title = {{Wave-Mamba: Wavelet} State Space Model for Ultra-High-Definition Low-Light Image Enhancement},
year = {2024},
doi = {10.1145/3664647.3681580},
booktitle = {Proceedings of the 32nd ACM International Conference on Multimedia},
pages = {1534–1543},
}

@inproceedings{akkaynak2018revised,
  title={A revised underwater image formation model},
  author={Akkaynak, Derya and Treibitz, Tali},
  booktitle={the IEEE/CVF Conference on Computer Vision and Pattern Recognition (CVPR)},
  pages={6723--6732},
  year={2018}
}

@inproceedings{Wang2025,
  author    = {H. Wang and N. Anantrasirichai and F. Zhang and D. Bull},
  title     = {{UW-GS: Distractor}-Aware 3D Gaussian Splatting for Enhanced Underwater Scene Reconstruction},
  booktitle = {Proceedings of the IEEE/CVF Winter Conference on Applications of Computer Vision (WACV)},
  year      = {2025},
}

@inproceedings{zhang2024atlantis,
  title={Atlantis: Enabling Underwater Depth Estimation with Stable Diffusion},
  author={Zhang, Fan and You, Shaodi and Li, Yu and Fu, Ying},
  booktitle={the IEEE/CVF Conference on Computer Vision and Pattern Recognition (CVPR)},
  pages={11852--11861},
  year={2024}
}

@article{dekker2001imaging,
  title={Imaging spectrometry of water},
  author={Dekker, Arnold G and Brando, Vittorio E and Anstee, Janet M and Pinnel, Nicole and Kutser, Tiit and Hoogenboom, Erin J and Peters, Steef and Pasterkamp, Reinold and Vos, Robert and Olbert, Carsten and others},
  journal={Imaging spectrometry: Basic principles and prospective applications},
  pages={307--359},
  year={2001},
  publisher={Springer}
}

@book{dutre2018advanced,
  title={Advanced global illumination},
  author={Dutre, Philip and Bekaert, Philippe and Bala, Kavita},
  year={2018},
  publisher={AK Peters/CRC Press}
}

@InProceedings{Xie_2024_CVPR,
    author    = {Xie, Yaofeng and Kong, Lingwei and Chen, Kai and Zheng, Ziqiang and Yu, Xiao and Yu, Zhibin and Zheng, Bing},
    title     = {{UVEB: A} Large-scale Benchmark and Baseline Towards Real-World Underwater Video Enhancement},
    booktitle = {Proceedings of the IEEE/CVF Conference on Computer Vision and Pattern Recognition (CVPR)},
    month     = {June},
    year      = {2024},
    pages     = {22358-22367}
}

@article{bai2024retinexmamba,
  title     = {{Retinexmamba: Retinex}-based Mamba for Low-light Image Enhancement},
  author    = {Jiesong Bai and Yuhao Yin and Qiyuan He and Yuanxian Li and Xiaofeng Zhang},
  journal   = {arXiv preprint},
  volume    = {arXiv:2405.03349},
  year      = {2024}
}

@inproceedings{Peng:FourierTMamba:2024,
author = {Peng, Shuwei and Zhang, Xu and Jiang, Aiwen and Liu, Changhong and Ye, Jihua},
title = {Low-Light Image Enhancement via FourierTMamba: A Hybrid Frequency-Spatial Approach},
year = {2024},
doi = {10.1145/3696409.3700175},
booktitle = {Proceedings of the 6th ACM International Conference on Multimedia in Asia}
}

@InProceedings{Yang:Implicit:2023,
    author    = {Yang, Shuzhou and Ding, Moxuan and Wu, Yanmin and Li, Zihan and Zhang, Jian},
    title     = {Implicit Neural Representation for Cooperative Low-light Image Enhancement},
    booktitle = {Proceedings of the IEEE/CVF International Conference on Computer Vision (ICCV)},
    month     = {October},
    year      = {2023},
    pages     = {12918-12927}
}

@InProceedings{Yi:Diff:2023,
    author    = {Yi, Xunpeng and Xu, Han and Zhang, Hao and Tang, Linfeng and Ma, Jiayi},
    title     = {Diff-Retinex: Rethinking Low-light Image Enhancement with A Generative Diffusion Model},
    booktitle = {Proceedings of the IEEE/CVF International Conference on Computer Vision (ICCV)},
    month     = {October},
    year      = {2023},
    pages     = {12302-12311}
}

@article{Lin:BVIRLV:2024,
  title={{BVI-RLV: A} Fully Registered Dataset and Benchmarks for Low-Light Video Enhancement},
  author={Ruirui Lin and Nantheera Anantrasirichai and Guoxi Huang and Joanne Lin and Qi Sun and Alexandra Malyugina and David R Bull},
  journal={arXiv preprint arXiv:2401.10166},
  year={2024}
}

@article{zhu2024vision,
  title={Vision mamba: Efficient visual representation learning with bidirectional state space model},
  author={Zhu, Lianghui and Liao, Bencheng and Zhang, Qian and Wang, Xinlong and Liu, Wenyu and Wang, Xinggang},
  journal={arXiv preprint arXiv:2401.09417},
  year={2024}
}

@article{olsson2022context,
  title={In-context learning and induction heads},
  author={Olsson, Catherine and Elhage, Nelson and Nanda, Neel and Joseph, Nicholas and DasSarma, Nova and Henighan, Tom and Mann, Ben and Askell, Amanda and Bai, Yuntao and Chen, Anna and others},
  journal={arXiv preprint arXiv:2209.11895},
  year={2022}
}

@article{blelloch1990prefix,
  title={Prefix sums and their applications},
  author={Blelloch, Guy E},
  year={1990},
  publisher={School of Computer Science, Carnegie Mellon University Pittsburgh, PA, USA}
}

@INPROCEEDINGS{Lin:CVMP:2024,  
author={Ruirui Lin, Qi Sun and Nantheera Anantrasirichai},  
booktitle={European Conference on Visual Media Production},   
title={Low-light Video Enhancement with Conditional Diffusion Models and Wavelet Interscale Attentions},   
year={2024}}

@ARTICLE{retinexDIP,
  author={Zhao, Zunjin and Xiong, Bangshu and Wang, Lei and Ou, Qiaofeng and Yu, Lei and Kuang, Fa},
  journal={IEEE Transactions on Circuits and Systems for Video Technology (TCSVT)}, 
  title={RetinexDIP: A Unified Deep Framework for Low-Light Image Enhancement}, 
  year={2022},
  volume={32},
  number={3},
  pages={1076-1088},
  doi={10.1109/TCSVT.2021.3073371}}

@ARTICLE{LECARM,
  author={Ren, Yurui and Ying, Zhenqiang and Li, Thomas H. and Li, Ge},
  journal={IEEE Transactions on Circuits and Systems for Video Technology}, 
  title={LECARM: Low-Light Image Enhancement Using the Camera Response Model}, 
  year={2019},
  volume={29},
  number={4},
  pages={968-981},
  doi={10.1109/TCSVT.2018.2828141}}
\bibliographystyle{spiebib} 

\end{document}